\pdfoutput=1

\documentclass[11pt]{article}

\usepackage[]{acl}
\usepackage{times}
\usepackage{latexsym}

\usepackage[T1]{fontenc}

\usepackage[utf8]{inputenc}

\usepackage{microtype}

\usepackage{algorithm}
\usepackage{algorithmic}
\usepackage{siunitx}    
\usepackage{relsize}
\usepackage{etoolbox}
\robustify\smaller

\usepackage{newfloat}
\usepackage{listings}
\floatstyle{ruled}
\newfloat{listing}{tb}{lst}{}
\floatname{listing}{Listing}

\usepackage{soul}
\usepackage{amsfonts}
\usepackage{multirow}

\usepackage{dsfont}
\usepackage{commath}

\usepackage{amsmath}
\usepackage{mathtools}

\usepackage{bbold}

\usepackage[utf8]{inputenc}
\usepackage{pgfplots}
\DeclareUnicodeCharacter{2212}{−}
\usepgfplotslibrary{groupplots,dateplot}
\usetikzlibrary{patterns,shapes.arrows}
\pgfplotsset{compat=newest}

\usepackage{placeins}
\usepackage{subcaption}
\usepackage{mwe}
\usepackage{pifont}
\usepackage{xcolor}
\usepackage{paralist}

\usepackage{booktabs}
\usepackage{relsize}
\usepackage{array}

\newcolumntype{V}{>{\smaller}l}

\newcommand{\class}[1]{\textsf{#1}\xspace}
\newcommand{\AAE}{\class{AAE}}
\newcommand{\SAE}{\class{SAE}}
\newcommand{\happy}{\class{happy}}
\newcommand{\sad}{\class{sad}}

\newcommand{\dataset}[1]{\textsc{#1}\xspace}
\newcommand{\Moji}{\dataset{Moji}}
\newcommand{\Bios}{\dataset{Bios}}

\newcommand{\method}[1]{\textsc{#1}\xspace}
\newcommand{\Standard}{\method{Standard}}
\newcommand{\INLP}{\method{INLP}}

\newcommand{\Adv}{\method{Adv}}
\newcommand{\DAdv}{\method{DAdv}}

\newcommand{\Disc}{\method{Discriminator}}
\newcommand{\FairBatch}{\method{FairBatch}}
\newcommand{\AAdv}{\method{A-Adv}}
\newcommand{\ADAdv}{\method{A-DAdv}}

\newcommand{\GAP}{\ensuremath{\text{GAP}}\xspace}
\newcommand{\DTO}{\ensuremath{\text{DTO}}\xspace}

\newcommand\hvec{\mathbf{h}}

\newcommand\gvec{\mathbf{g}}
\newcommand\yvec{\mathbf{y}}
\newcommand\xvec{\mathbf{x}}

\usepackage{bm}
\newcommand\thetavec{\bm{\theta}}
\newcommand\phivec{\bm{\phi}}

\usepackage{xspace}
\usepackage{adjustbox}

\definecolor{myBlue}{rgb}{0.12156863, 0.46666667, 0.70588235}
\definecolor{myOrange}{rgb}{1., 0.49803922, 0.05490196}
\definecolor{myRed}{rgb}{1,0,0}

\usepackage{mathtools, caption}

\usepackage{circuitikz}

\title{Towards Equal Opportunity Fairness through Adversarial Learning
}

\author{
Xudong Han$^{1}$ \qquad  Timothy Baldwin$^{1,2}$ \qquad    Trevor Cohn$^{1}$\\ 
$^{1}$The University of Melbourne \\ 
$^{2}$MBZUAI \\ 
\url{xudongh1@student.unimelb.edu.au}, \url{{tbaldwin,t.cohn}@unimelb.edu.au}
}

\begin{document}
\maketitle
\begin{abstract}
Adversarial training is a common approach for bias mitigation in natural language processing.
Although most work on debiasing is motivated by equal opportunity, it is not explicitly captured in standard adversarial training.
In this paper, we propose an augmented discriminator for adversarial training, which takes the target class as input to create richer features and more explicitly model equal opportunity.
Experimental results over two datasets show that our method substantially improves over standard adversarial debiasing methods, in terms of the performance--fairness trade-off.

\end{abstract}

\section{Introduction}

While natural language processing models have achieved great successes across a variety of classification tasks in recent years, naively-trained models often learn spurious correlations with confounds like user demographics and socio-economic factors~\citep{badjatiya2019stereotypical, zhao2018gender, li-etal-2018-towards}.

A common way of mitigating bias relies on ``unlearning'' discriminators during the debiasing process.
For example, in adversarial training, an encoder and discriminator are trained such that the encoder attempts to prevent the discriminator from identifying protected attributes~\citep{zhang2018mitigating,li-etal-2018-towards,han2021diverse}.
Intuitively, such debiased representations from different protected groups are not separable, and as a result, any classifier applied to debiased representations will also be agnostic to the protected attributes.
The fairness metric corresponding to such debiasing methods is known as \emph{demographic parity}~\citep{feldman2015certifying}, which is satisfied if classifiers are independent of the protected attribute.
Taking loan applications as an example, \emph{demographic parity} is satisfied if candidates from different groups have the same approval rate.

However, demographic parity has its limitations, as illustrated by~\citet{barocas-hardt-narayanan}: a hirer that carefully selects applicants from group $a$ and arbitrarily selects applicants from group $b$ with the same acceptance rate of $p>0$ achieves demographic parity, but is far from fair as they are more likely to select inappropriate applicants in group $b$.

In acknowledgement of this issue, \citep{hardt2016equality} proposed: (1) \emph{equalized odds}, which is satisfied if a classifier is independent of protected attributes within each class, i.e., class-specific independent; and (2) \emph{equal opportunity}, which is a relaxed version of \emph{equalized odds} that only focuses on independence in ``advantaged'' target classes (such as the approval class for loans). 

\begin{figure}[t]
    \centering
    \includegraphics[width=0.45\textwidth]{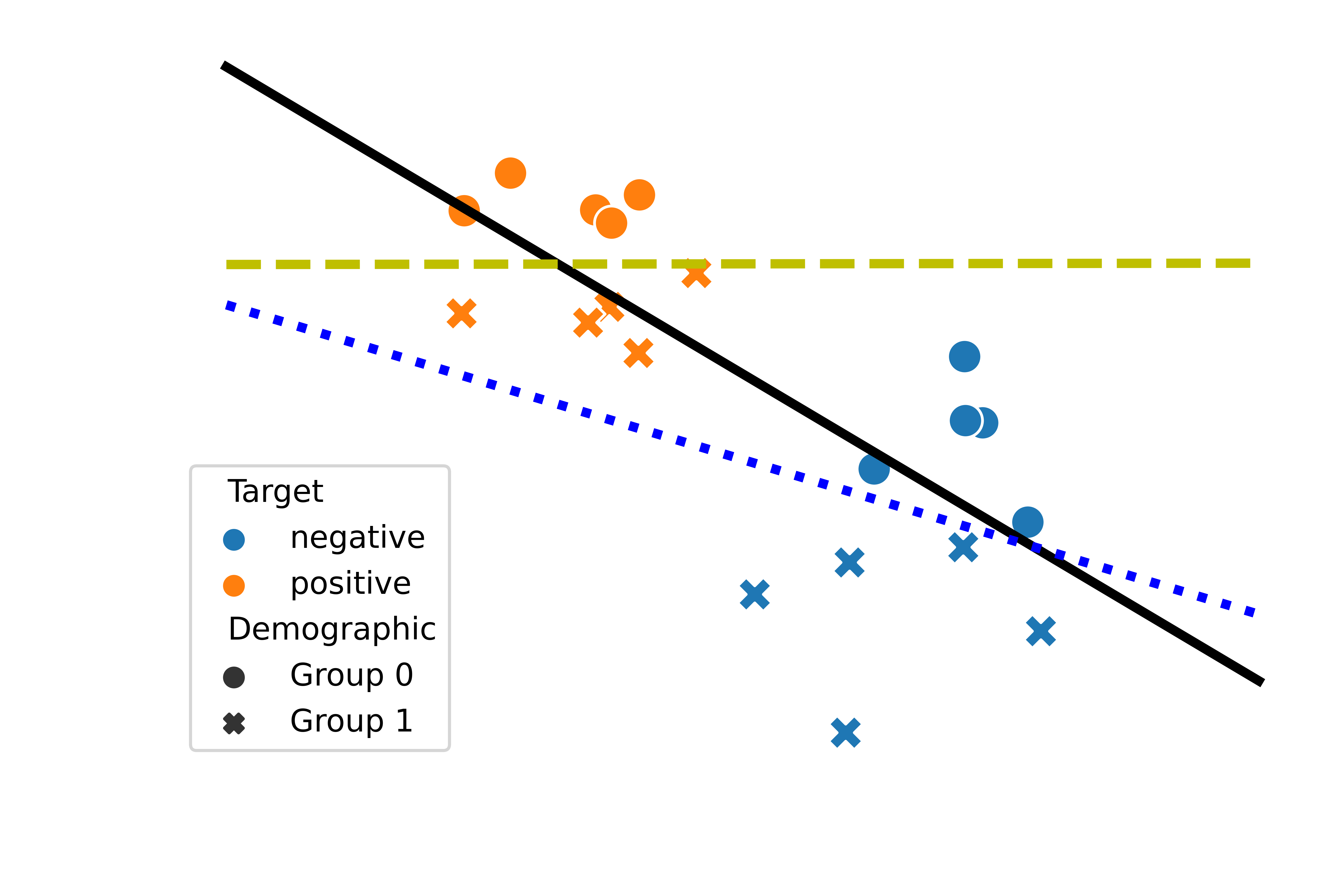}
    \caption{The black solid, yellow dashed, and blue dotted lines are the decision boundaries of linear discriminators for demographic trained over all instances, $\yvec=$~positive, and $\yvec=$~negative, resp. The decision boundaries learned by our proposed class-specific discriminators (yellow ashed and blue dotted) are distinct to the previous discriminator setting (black solid).}
    \label{fig:intuation}
\end{figure}

Similar to \emph{demographic parity}, standard adversarial training does not consider the target label for protected information removal, which is fundamental to equal opportunity.
Figure~\ref{fig:intuation} shows a toy example where hidden representations are labelled with the associated target labels via colour, and protected labels via shape. 
Taking the target label information into account and training separate discriminators for each of the two protected attributes, it can be seen that the linear decision boundaries are quite distinct, and each is different from the decision boundary when the target class is not taken into consideration.

To enable adversarial training to recognize the correlation between protected attributes and target classes, we propose a novel discriminator architecture that captures the class-specific protected attributes during adversarial training. 
Moreover, our proposed mechanism is generalizable to other SOTA variants of adversarial training, such as DAdv~\citep{han2021diverse}.
Experiments over two datasets show that our method consistently outperforms standard adversarial learning.

The source code has been included in \texttt{fairlib}~\citep{han2022fairlib}, and is available at \url{https://github.com/HanXudong/fairlib}.

\section{Related Work}
\subsection{Fairness Criterion}
Various types of fairness criteria have been proposed, which \citet{barocas-hardt-narayanan} divide into three categories according to the levels of (conditional) independence between protected attributes, target classes, and model predictions.

\textbf{Independence}, also known as \emph{demographic parity}~\citep{feldman2015certifying}, is satisfied iff predictions are not correlated with protected attributes. 
In practice, this requires that the proportion of positive predictions is the same for each protected group. 
One undesirable property of \emph{independence} is that base rates for each class must be the same across different protected groups, which is usually not the case~\citep{hardt2016equality}. 

\textbf{Separation}, also known as \emph{equalized odds}~\citep{hardt2016equality}, is satisfied iff predictions are independent of protected attributes conditioned on the target classes. 
In the case of \emph{equal opportunity}, this requirement is only applied to the positive (or advantaged) class.
In the multi-class classification setting, equal opportunity is applied to all classes, and becomes equivalent to equalized odds.

\textbf{Sufficiency}, also known as \emph{test fairness}~\citep{chouldechova2017fair}, is satisfied if the target class is independent of the protected attributes conditioned on the predictions. 

In this paper, we focus on the separation criterion, and employ the generalized definition of \emph{equal opportunity} for both binary and multi-class classification.

\subsection{Bias Mitigation}
Here we briefly review bias mitigation methods that can be applied to a \Standard model (i.e., do not involve any adjustment of the primary objective function to account for bias).

\textbf{Pre-processing} manipulates the input distribution before training, to counter for data bias.
Typical methods include dataset augmentation~\citep{zhao2018gender}, dataset resampling~\citep{wang2019balanced}, and instance reweighting~\citep{NEURIPS2020_07fc15c9}.
One limitation of these methods is that the distribution adjustment is statically determined based on the training set distribution, and not updated later.
To avoid this limitation, \citet{Roh2021FairBatch} proposed \FairBatch to adjust the input distributions dynamically. 
Specifically, \FairBatch formulates the model training as a bi-level optimization problem: the inner optimizer is the same as \Standard, while the outer optimizer adjusts the resampling probabilities of instances based on the loss difference between different demographic groups and target classes from the inner optimization. 

\textbf{At-training-time} introduces constraints into the optimization process for model training.
We mainly focus on the adversarial training (\Adv), which jointly trains a \Standard model and a \Disc component to alleviate protected attributes~\citet{elazar2018adversarial,li-etal-2018-towards}.
More recently, \citet{han2021diverse} present \DAdv, which employs multiple discriminators with orthogonality constraints to unlearn protected information.

\textbf{Post-processing} aims at adjusting a trained classifier based on protected attributes, such that the final predictions are fair to the different protected groups. 
A popular method is \INLP~\citep{ravfogel-etal-2020-null}, which trains a \Standard model, and then iteratively projects the last-hidden-layer's representations to the null-space of demographic information, and uses the projected hidden representations to make predictions.

\subsection{Model Comparison}
In contrast to single-objective evaluation, evaluation of fairness approaches generally considers fairness and performance simultaneously. 
Typically, no single method simultaneously achieves the best performance and fairness, making comparison difficult. 

\begin{figure}[t!]
    \centering
     \begin{subfigure}[b]{0.45\textwidth}
         \centering
         \includegraphics[width=\linewidth]{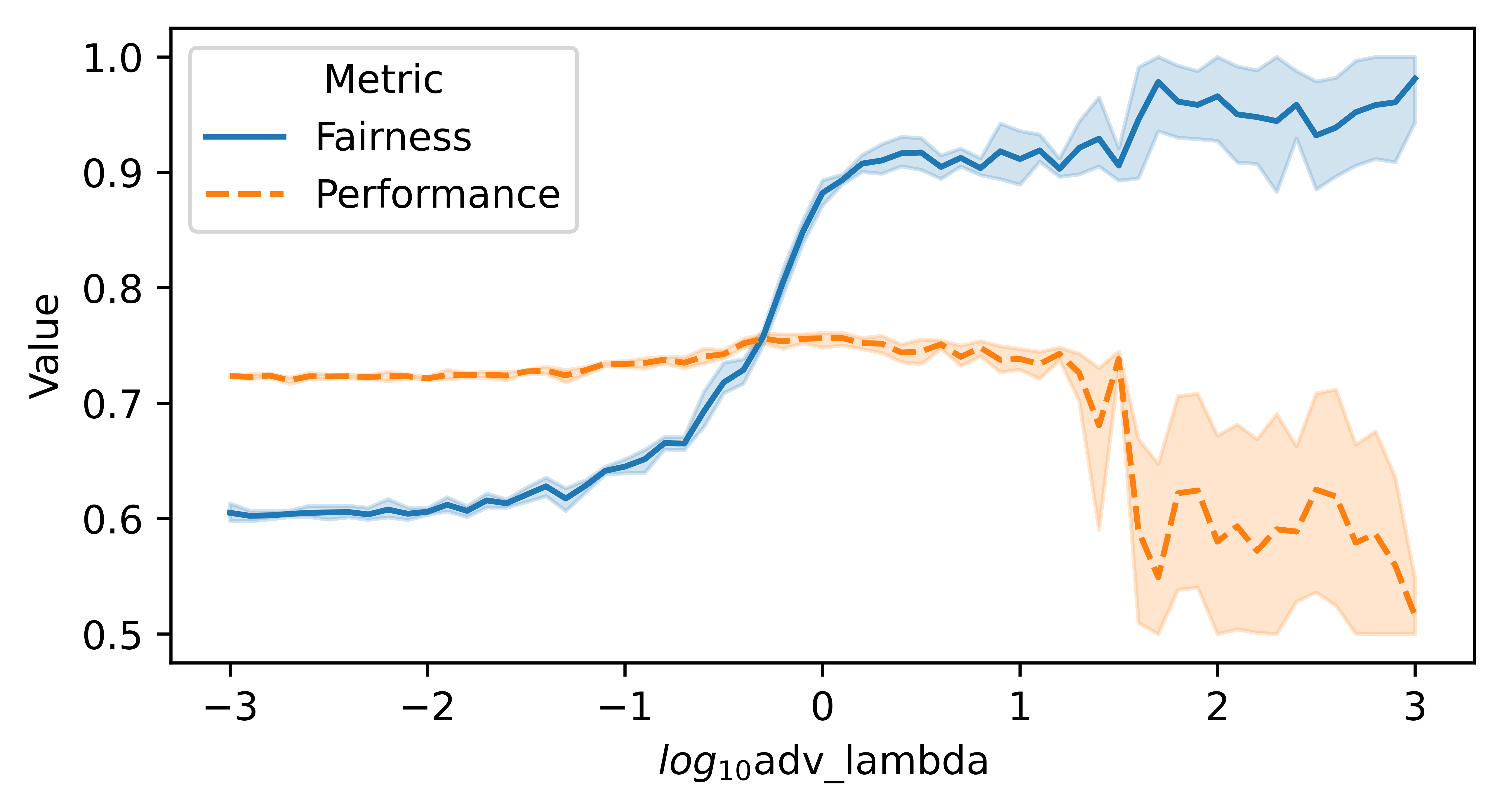}
         \caption{Tuning \Adv trade-off hyperparameter}
         \label{fig:moji_adv_hypertune}
    \end{subfigure}
    \hfill 
    \begin{subfigure}[b]{0.45\textwidth}
        \centering
        \includegraphics[width=\linewidth]{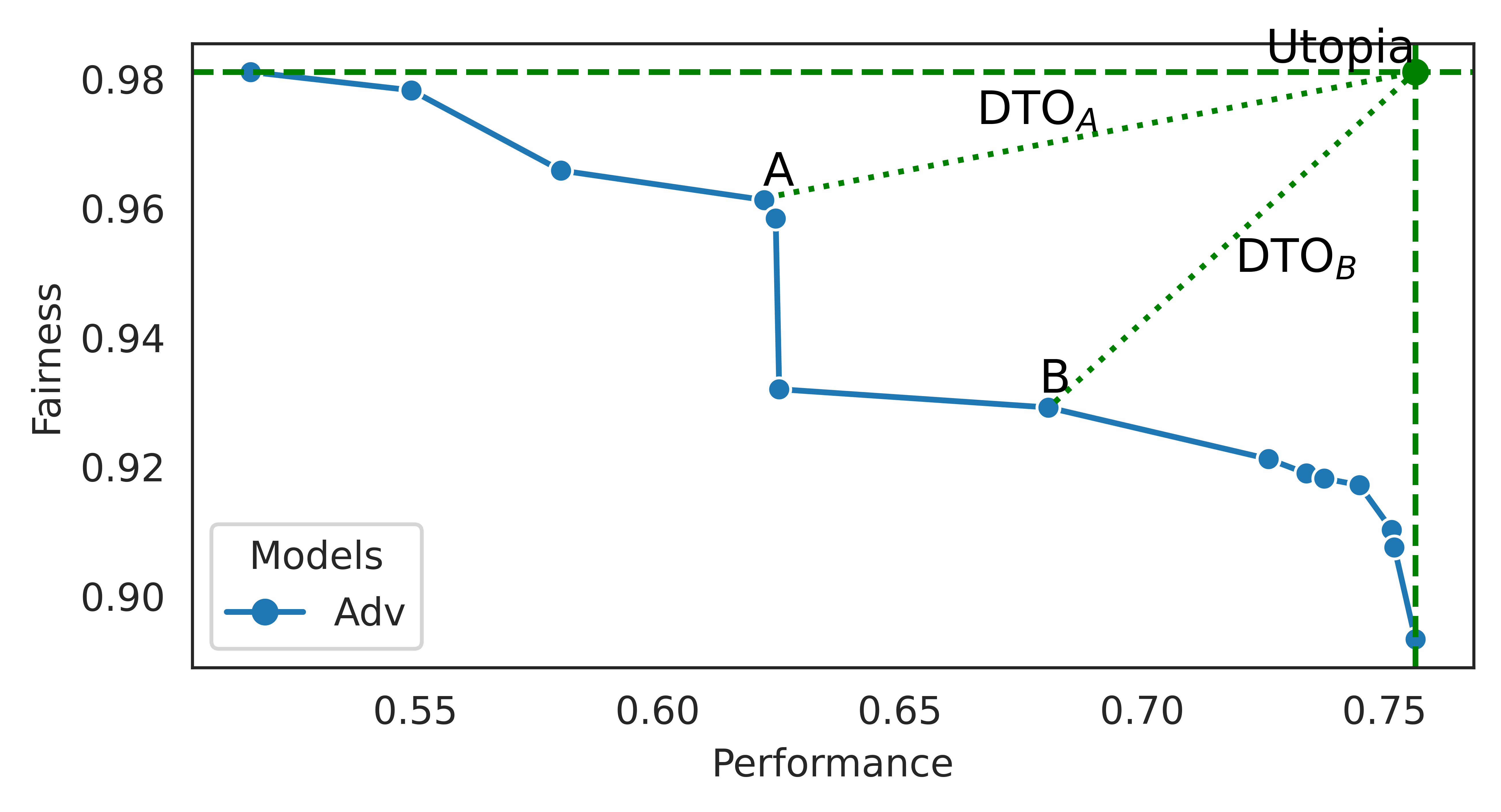}
        \caption{\Adv trade-off}
        \label{fig:moji_adv_tradeoff}
    \end{subfigure}
    \hfill
    \caption{An example of performance--fairness trade-off with respect to different values of the strength for the additional loss for discriminator unlearning in \Adv. The shaded area refers to 95\% CI. Figure~\ref{fig:moji_adv_tradeoff} also provides an example for \DTO. The green dashed vertical and horizontal lines denote the best performance and fairness, respectively, and their intersection point is the Utopia point. The length of the green dotted lines from A and B to the Utopia point are the \DTO for candidate models A and B, respectively.}
    \label{fig:adv_tune}
\end{figure}

\textbf{Performance--Fairness Trade-off} is a common way of comparing different debiasing methods without the requirement for model selection. 
Specifically, debiasing methods typically involve a trade-off hyperparameter, which controls the extent to which the final model sacrifices performance for fairness, such as the strength of discriminator unlearning in \Adv, as shown in Figure~\ref{fig:moji_adv_hypertune}. 

Typically, instead of looking at the performance/fairness for different trade-off hyperparameter values, it is more meaningful to focus on the Pareto plots (Figure~\ref{fig:moji_adv_tradeoff}), which show the maximum fairness that can be achieved at different performance levels, and vice versa.

\textbf{\DTO} is a metric to quantify the performance--fairness tradeoff, which measures the \textbf{D}istance \textbf{T}o the \textbf{O}ptimal point for candidate models~\citep{salukvadze1971concerning, marler2004survey, han2021balancing}. 
Note that the \DTO can not only be used to compare different methods, but also for early stopping and model selection.
Figure~\ref{fig:moji_adv_tradeoff} provides an example of model selection based on \DTO, where the optimal (Utopia) point at the top-right corner and \DTO scores are the normalized Euclidean distance (length of green dotted lines) between the optimal points and candidate models.

In this paper, we use \DTO for early stopping and method comparison, while selecting the best model for each method based on the results of \Standard.

\section{Methods}

\begin{figure}
\centering

\usetikzlibrary{decorations,calc,decorations.pathmorphing,decorations.pathreplacing}

\begin{subfigure}[t]{\linewidth}
\centering

\tikzstyle{block} = [draw, fill=blue!20, rectangle, rounded corners,
    minimum height=0.75cm, minimum width=0.75cm]
\tikzstyle{sum} = [draw, fill=green!20, circle, node distance=2cm]
\tikzstyle{input} = [coordinate]
\tikzstyle{output} = [coordinate]
\tikzstyle{pinstyle} = [pin edge={to-,thin,black}]
\tikzstyle{ArrowC1} = [rounded corners, thick]
\begin{tikzpicture}[auto, node distance=1cm,>=latex, scale = 0.8, transform shape]
    \node [block, fill=green!20, name=input] {$m$};
    \node [input, left of=input, node distance=1cm] (input_x) {};
    \node [input, right of=input, node distance=1.5cm] (hidden) {};
    \node [block, right of=hidden,fill=green!20, node distance=1.75cm] (MC) {$f$};
    \node [input, above of=MC, node distance=1.5cm] (adv1) {};
    
    \node [input, above of=input_x, node distance=1.6cm] (input_y) {};
    \node [input, above of=input, node distance=1.6cm] (input_y2) {};
    
    \node [block, right of=adv1, node distance=2cm] (dense1) {$d$};

    \node [output, right of=dense1, node distance=1.5cm] (out1) {};
    \node [output, below of=out1, node distance=1.5cm] (out0) {};
    
    \draw [-] (input) -- node[] {$\hvec$} (hidden);
    
    \draw [->] (input_x) -- node[] {$\xvec$} (input);
    \draw[->] (hidden) -- (MC);
    \draw[-, dashed, rounded corners=.15cm] (hidden.south) |- ($(adv1.west)+(0, 0)$);
    
    \draw [->, dashed] (adv1) -- (dense1);
    
    \draw [->] (dense1) -- node[] {$\hat{\gvec}$} (out1);
    \draw [->] (MC) -- node[] {$\hat{\yvec}$} (out0);
\end{tikzpicture}
\caption{Standard Adversarial}
\label{fig:stand_Adv}
\end{subfigure} 
\\
\begin{subfigure}[t]{\linewidth}
\centering

\tikzstyle{block} = [draw, fill=blue!20, rectangle, rounded corners,
    minimum height=0.75cm, minimum width=0.75cm]
\tikzstyle{sum} = [draw, fill=green!20, circle, node distance=2cm]
\tikzstyle{input} = [coordinate]
\tikzstyle{output} = [coordinate]
\tikzstyle{pinstyle} = [pin edge={to-,thin,black}]
\tikzstyle{ArrowC1} = [rounded corners, thick]
\begin{tikzpicture}[auto, node distance=1cm,>=latex, scale = 0.8, transform shape]
    \node [block, fill=green!20, name=input] {$m$};
    \node [input, left of=input, node distance=1cm] (input_x) {};
    \node [input, right of=input, node distance=1.5cm] (hidden) {};
    \node [block, right of=hidden,fill=green!20, node distance=1.75cm] (MC) {$f$};
    \node [block, above of=MC, node distance=1.5cm] (adv1) {$a$};
    
    \node [input, above of=input_x, node distance=1.6cm] (input_y) {};
    \node [input, above of=input, node distance=1.6cm] (input_y2) {};
    
    \node [block, right of=adv1, node distance=2cm] (dense1) {$d$};

    \node [output, right of=dense1, node distance=1.5cm] (out1) {};
    \node [output, below of=out1, node distance=1.5cm] (out0) {};
    
    \draw [-] (input) -- node[] {$\hvec$} (hidden);
    
    \draw [->] (input_y) -- node[] {$\yvec$} ($(adv1.west)+(0, 0.1)$);
    \draw [->] (input_x) -- node[] {$\xvec$} (input);
    \draw[->] (hidden) -- (MC);
    \draw[->, dashed, rounded corners=.15cm] (hidden.south) |- ($(adv1.west)+(0, -0.1)$);
    
    \draw [->] (adv1) -- node[] {$\hvec^{a}$} (dense1);
    
    \draw [->] (dense1) -- node[] {$\hat{\gvec}$} (out1);
    \draw [->] (MC) -- node[] {$\hat{\yvec}$} (out0);
\end{tikzpicture}
\caption{Augmented Adversarial}
\label{fig:augmented_adv}
\end{subfigure}
\\
\begin{subfigure}[t]{\linewidth}
\centering

\tikzstyle{block} = [draw, fill=blue!20, rectangle, rounded corners,
    minimum height=0.75cm, minimum width=1.2cm]
\tikzstyle{roundnode}={circle, draw=red!60, fill=red!20,minimum size=7mm},
\tikzstyle{sum} = [draw, fill=green!20, circle]
\tikzstyle{input} = [coordinate]
\tikzstyle{output} = [coordinate]
\tikzstyle{pinstyle} = [pin edge={to-,thin,black}]
\tikzstyle{ArrowC1} = [rounded corners, thick]
\begin{tikzpicture}[auto, node distance=1cm,>=latex, scale = 0.8, transform shape]

    \node [block, name=ms] {$m^{s}$};
    
    \node [input, left of=ms, node distance=1.5cm] (input_hs) {};
    \draw[->] (input_hs) -- (ms);
    
    \node [input, left of=input_hs, node distance=1cm] (input_h) {};
    \draw[-] (input_h) -- node[] {$\hvec$} (input_hs);
    
    \node [block, above of=ms, node distance=1.1cm] (mpc) {$m'_{|C|}$};
    \draw[->, rounded corners=.15cm] (input_hs.north) |- ($(mpc.west)+(0, -0)$);
    
    \node [block, above of=mpc, node distance=1.5cm] (mp1) {$m'_{1}$};
    \draw[->, rounded corners=.15cm] (input_hs.north) |- ($(mp1.west)+(0, -0)$);
    \path (mp1.south)+(0.0,-0.25) node [] {$\vdots$}; 
    
    \node [input, right of=mp1, node distance=1cm] (temp1) {};
    \node [input, below of=temp1, node distance=0.75cm] (temp2) {};
    \draw[-, rounded corners=.15cm] (mp1.east) -|(temp2.north);
    \draw[-, rounded corners=.15cm] (mpc.east) -|(temp2.south);
    
    \node [roundnode, right of=temp2, node distance=1cm, fill=red!20] (stack) {stack};
    \draw[->] (temp2) -- (stack);
    
    \node[roundnode, right of=stack, node distance = 1.5cm, fill=red!20] (product) {$\times$};
    \draw[->] (stack) -- (product);
    
    \node [input, above of=input_h, node distance=3.2cm] (input_y) {};
    \draw[->, rounded corners=.15cm] (input_y.east) -| node[] {$\yvec$} (product.north);
    
    \node[roundnode, below of=product, node distance = 1.85cm, fill=red!20] (sum) {$+$};
    \draw[->] (product) -- (sum);
    \draw[->] (ms) -- (sum);
    
    \node [output, right of=sum, node distance=1.5cm] (out) {};
    \draw [->] (sum) -- node[] {$\hvec^{a}$} (out);
    
\end{tikzpicture}

\caption{Augmentation layer}
\label{fig:augmentation_layer}
\end{subfigure}

\label{fig:architecture}
\caption{
Proposed model architectures. Dashed lines denote gradient reversal in adversarial learning.
Green and blue rounded rectangles are the trainable neural network layers for target label classification and bias mitigation, resp. Red circles are operations.
}
\end{figure}
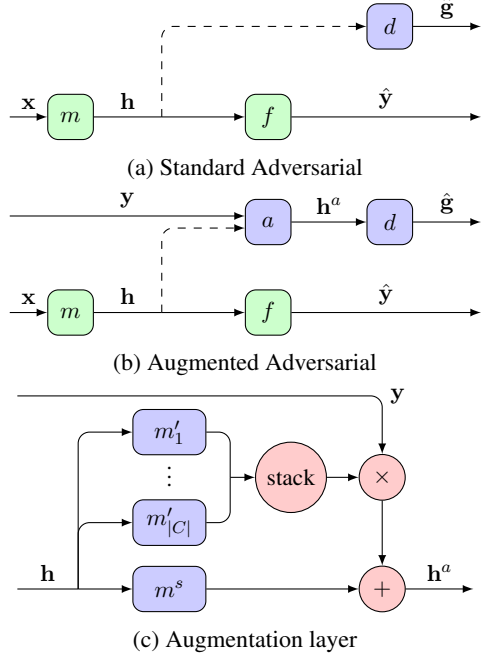

Here we describe the methods employed in this paper. 
Formally, as shown in Figure~\ref{fig:stand_Adv}, given an input $\xvec$ annotated with main task label $\yvec$ and protected attribute label $\gvec$, a main task model consists of two connected parts: the encoder $\hvec = m(\xvec; \thetavec^{m})$ is trained to compute the hidden representation from an input $\xvec$, and the classifier makes prediction, $\hat{\yvec}=f(\hvec; \thetavec^{f})$. 
During training, a discriminator $d$, parameterized by $\phivec^{d}$, is trained to predict $\hat{\gvec}=d(\hvec; \phivec^{d})$ from the final hidden-layer representation $\hvec$. 
The discriminator is only used as a regularizer during training, and will be dropped at the test time, i.e., the final model with adversarial training is the same as a naively trained model at the inference time.


\subsection{Adversarial Learning}
\label{sec:adv}

Following the setup of \citet{li-etal-2018-towards, han2021diverse}, the optimisation objective for standard adversarial training is:
\begin{equation}
\label{eq:objective}
    \min_{\thetavec^{m},\thetavec^{f}}\max_{\phivec^{*}}\mathcal{X}(\yvec, \hat{\yvec}) - \lambda \mathcal{X}(\gvec, \hat{\gvec})
\end{equation}
where $\phivec^{*}=\{\phivec^{d}\}$, $\mathcal{X}$ is the cross entropy loss, and $\lambda$ is a trade-off hyperparameter. 
Solving this minimax optimization problem encourages the main task model hidden representation $\hvec$ to be informative to $f$ and to be uninformative to $d$.

\subsection{Discriminator with Augmented Representation}
\label{sec:augmented}

As illustrated in Figure~\ref{fig:augmented_adv}, we propose \textbf{augmented discrimination}, a novel means of strengthening the adversarial component. 
Specifically, an extra augmentation layer $a$ is added between $m$ and $d$, where $a$ takes the $\yvec$ into consideration to create richer features, i.e., $\hat{\gvec} = d(a(\hvec;\yvec;\phivec^{a});\phivec^{d})$.


\paragraph{Augmentation Layer}
Figure~\ref{fig:augmentation_layer} shows the architecture of the proposed augmentation layer.
Inspired by the domain-conditional model of~\citet{li-etal-2018-whats}, the augmentation layer $a$ consists of one shared projector and $|C|$ specific projectors,$\{m^{s},{m'}_{1},{m'}_{2},\dots,{m'}_{|C|}\}$, where $|C|$ is the number of target classes.

Formally, let $m^{s}(\hvec; \phivec^{s})$ be a function parameterized by $\phivec^{s}$ which projects a hidden representation $\hvec$ to $\hvec^{s}$ representing features w.r.t $\gvec$ that are \emph{shared} across classes, and ${m'}_{j}(\hvec; \phivec^{j})$ be a class-specific function to the $j$-th class which projects the same hidden representation $\hvec$ to ${\hvec'}^{j}$ capturing features that are \emph{private} to the $j$-th class.
In this paper, we employ the same architecture for shared and all private projectors.
The resulting output of the augmentation layer is 
$$\hvec^{a} = a(\hvec;\yvec;\phivec^{a})=\hvec^{s}+\sum_{j=1}^{|C|}\yvec_{i,j}{\hvec'}^{j},$$
where $\phivec^{a} = \{\phivec^{s},\phivec^{1},\dots,\phivec^{|C|}\}$, and $\yvec_{i,:}$ is 1-hot.
Moreover, let $\phivec^{*}=\{\phivec^{d},\phivec^{a}\}$, the training objective is the same as Equation~\ref{eq:objective}.

Intuitively, $d$ is able to make better predictions over $\gvec$ based on $\hvec^{a}$ than the vanilla $\hvec$ due to the enhanced representations provided by $a$.
More formally, as the augmented discriminator models the conditional probability $\Pr(g|h,y)$, the unlearning of the augmented discriminator encourages conditional independence $\hvec \perp \gvec | \yvec$, which corresponds directly to the equal opportunity criterion.





\section{Experiments}
In order to compare our method with previous work, we follow the experimental setting of~\citet{han2021diverse}.
We provide full experimental details in Appendix~\ref{sec:Reproducibility}.

\subsection{Evaluation Metrics}
Following \citet{han2021diverse, ravfogel-etal-2020-null}, we use overall accuracy as the performance metric, and measure TPR \GAP for equal opportunity fairness.
For multiclass classification tasks, we report the quadratic mean (RMS) of TPR \GAP over all classes. 
While in a binary classification setup, TPR and TNR are equivalent to the TPR of the positive and negative classes, respectively, so we employ the RMS TPR \GAP in this case also.
For \GAP metrics, the smaller, the better, and a perfectly fair model will achieve 0 \GAP.
We further measure the fairness as 1-\GAP, the larger, the better.

More specifically, the calculation of RMS TPR \GAP consists of aggregations at the group and class levels.
At the group level, we measure the absolute TPR difference of each class between each group and the overall TPR $GAP^{TPR}_{G,y}=\sum_{g\in{G}}|TPR_{g,y}-TPR_{y}|$, and at the next level, we further perform the RMS aggregation at the class level to get the RMS TPR \GAP as $GAP = \sqrt{\frac{1}{|Y|}\sum_{y\in{Y}}(GAP^{TPR}_{G,y})^{2}}$. 

\begin{figure*}[ht!]
    \centering
    \begin{subfigure}[b]{0.48\textwidth}
        \centering
        \includegraphics[width=\textwidth]{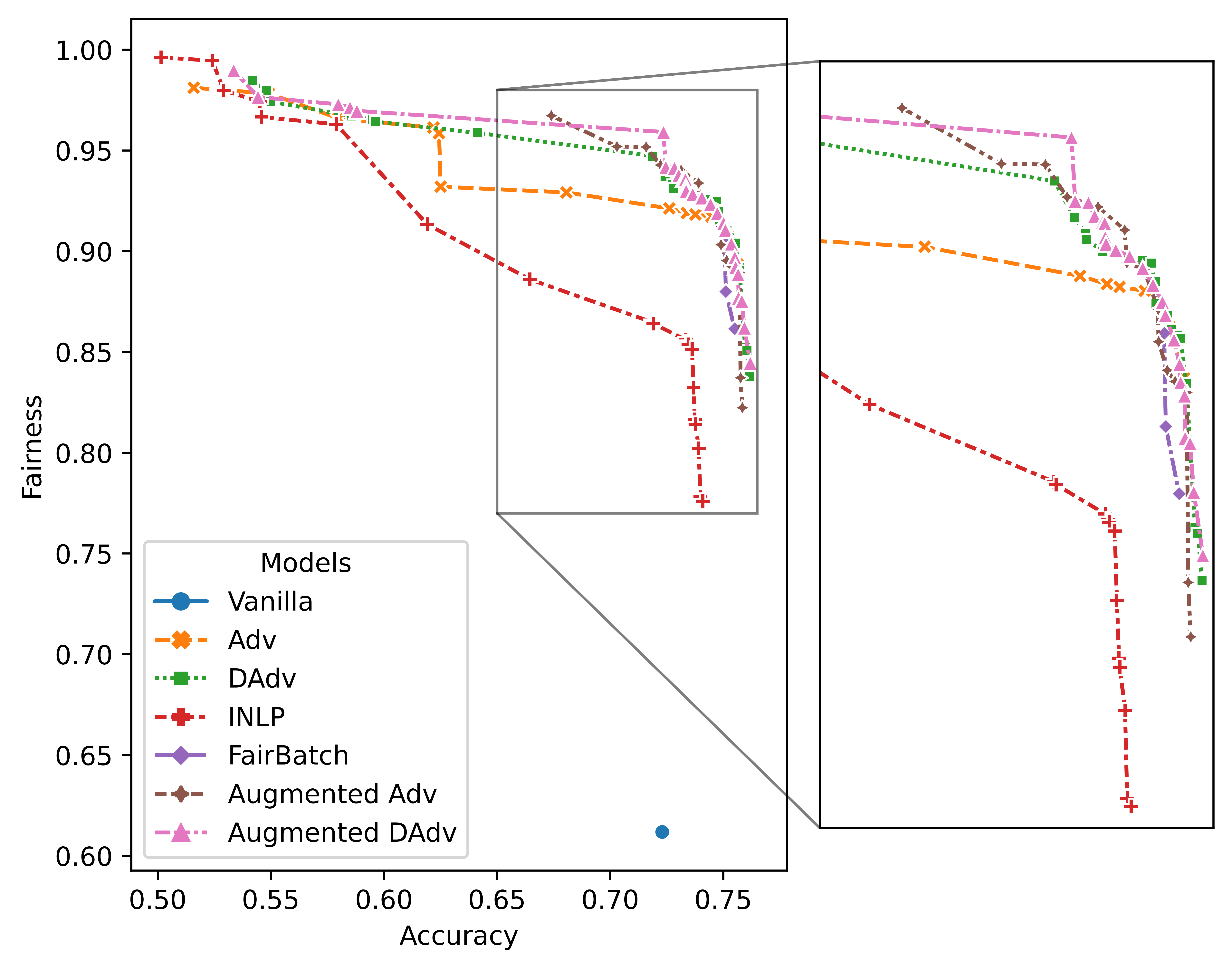}
        \caption{\Moji}
        \label{fig:adv_moji}
    \end{subfigure}
    \hfill 
    \begin{subfigure}[b]{0.48\textwidth}
        \centering
        \includegraphics[width=\textwidth]{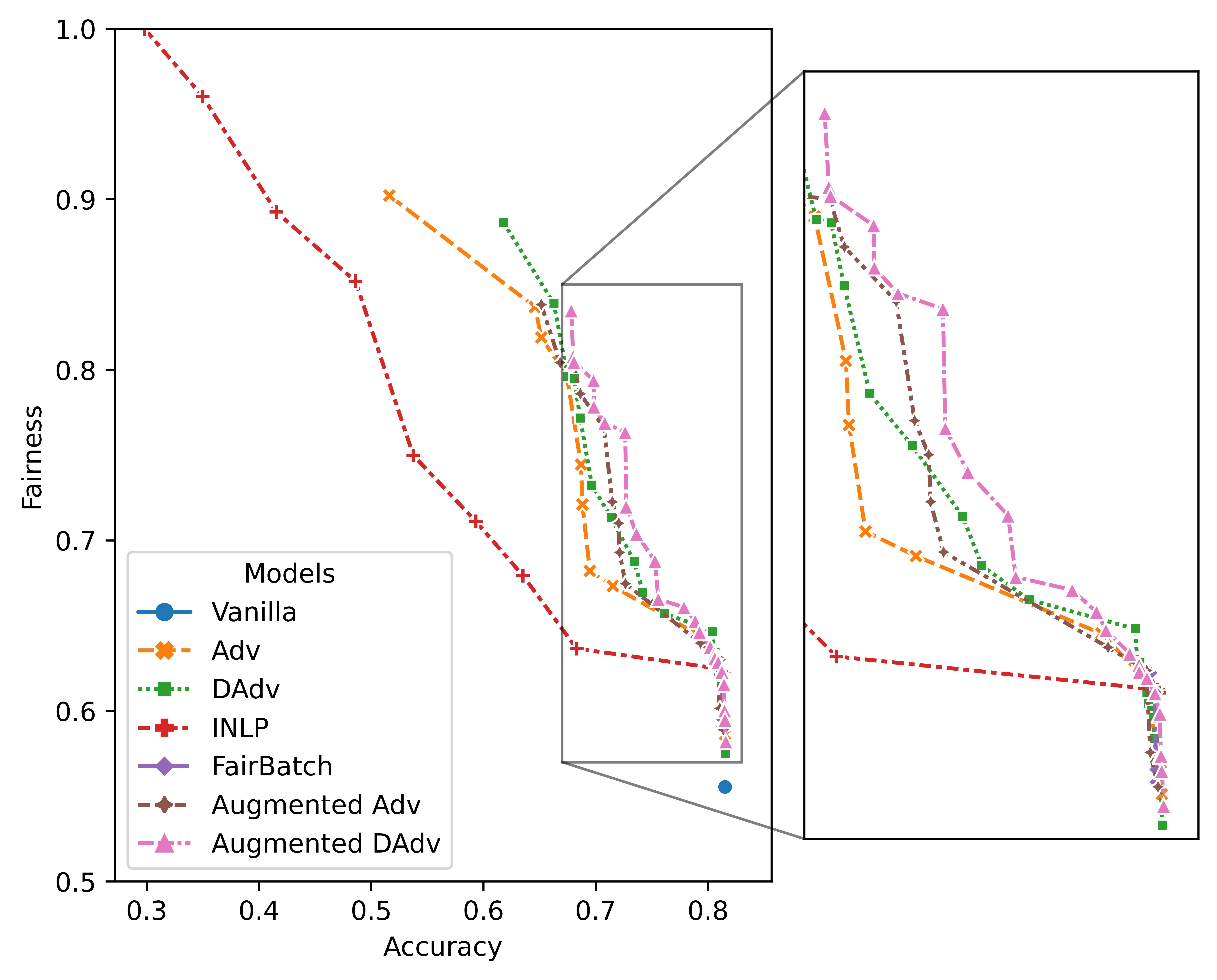}
        \caption{\Bios}
        \label{fig:adv_bios}
    \end{subfigure}
    \caption{Pareto frontiers of trade-offs between performance and fairness. The top-right represents ideal model with the idea performance and fairness.
    }
    \label{fig:tradeoffs}
\end{figure*}

\subsection{Dataset}
Following \citet{subramanian-etal-2021-evaluating}, we conduct experiments over two NLP classification tasks --- sentiment analysis and biography classification --- using the same dataset splits as prior work. 

\paragraph{\Moji}
This sentiment analysis dataset was collected by \citet{blodgett-etal-2016-demographic}, and contains tweets that are either African American English (AAE)-like or Standard American English (SAE)-like.
Each tweet is annotated with a binary `race' label (based on language use: either \AAE or \SAE) and a binary sentiment score determined by (redacted) emoji contained in it.

\paragraph{\Bios} 
The second task is biography classification~\citep{de2019bias, ravfogel-etal-2020-null}, where biographies were scraped from the web, and annotated for the protected attribute of binary gender and target label of 28 profession classes.

Besides the binary gender attribute, we additionally consider economic status as a second protected attribute. \citet{subramanian-etal-2021-evaluating} semi-automatically label economic status (wealthy vs.\ rest) based on the country the individual is based in, as geotagged from the first sentence of the biography.
For bias evaluation and mitigation, we consider the intersectional groups, i.e., the Cartesian product of the two protected attributes, leading to 4 intersectional classes: female--wealthy, female--rest, male--wealthy, and male--rest.

\subsection{Models}
\label{sec:models}

We first implement a naively trained model on each dataset, without explicit debiasing. On the \Moji dataset, we use DeepMoji~\citep{felbo2017} as the fixed encoders to get 2304d representations of input texts. 
For the \Bios dataset, we use uncased BERT-base~\citep{devlin2019bert}, taking the `AVG' representations extracted from the pretrained model, without further fine-tuning.

For adversarial method, both the \Adv and augmented \Adv, we jointly train the discriminator and classifier. 
Again, we follow~\citet{han2021diverse} in using a non-linear discriminator, which is implemented as a trainable 3-layer MLP.

A common problem is that a large number of instances are not annotated with protected attributes, e.g.\ only 28\% instances in the \Bios dataset are annotated with both gender and economic status labels.
The standard adversarial method has required all training instances are annotated with protected attributes, and thus can only be trained over a full-labelled subset, decreasing the training set size significantly.
To maintain the performance of the debiased model, we follow \citet{han-etal-2021-decoupling} in decoupling the training of the model and the discriminator, making it possible to use all instances for model training at a cost of the performance-fairness trade-off.

\subsection{Main results}

Figure~\ref{fig:tradeoffs} shows the results.
Each point denotes a candidate model, and we take the average over 5 runs with different random seeds.

\begin{table*}[ht!]

\begin{subfigure}{\textwidth}
\renewrobustcmd{\bfseries}{\fontseries{b}\selectfont}
\centering
\begin{adjustbox}{max width=\linewidth}
\sisetup{
round-mode = places,
}%
\begin{tabular}{
l
S[table-format=2.2, round-precision = 2]@{\,\( \pm \)\,}S[table-format=2.2, round-precision = 2,table-number-alignment = left]
S[table-format=2.2, round-precision = 2]@{\,\( \pm \)\,}S[table-format=2.2, round-precision = 2,table-number-alignment = left]
S[table-format=2.2, round-precision = 2]
S[table-format=2.2, round-precision = 2]@{\,\( \pm \)\,}S[table-format=2.2, round-precision = 2,table-number-alignment = left]
S[table-format=2.2, round-precision = 2]@{\,\( \pm \)\,}S[table-format=2.2, round-precision = 2,table-number-alignment = left]
S[table-format=2.2, round-precision = 2]
}

\toprule
& \multicolumn{5}{c}{\bf Trade-off 5\%} & \multicolumn{5}{c}{\bf Trade-off 10\%} \\
\cmidrule(lr){2-6}\cmidrule(lr){7-11}

\bf Model         & \multicolumn{2}{c}{\bf Accuracy$\uparrow$}     &\multicolumn{2}{c}{\bf Fairness$\uparrow$} & \multicolumn{1}{c}{\bf \DTO$\downarrow$ } & \multicolumn{2}{c}{\bf Accuracy$\uparrow$}     &\multicolumn{2}{c}{\bf Fairness$\uparrow$} & \multicolumn{1}{c}{\bf \DTO$\downarrow$ } \\ 
\midrule

\Standard   &            72.2981 &              00.4576 &            61.1870 &           01.4356 & 47.6849 & 72.2981 &              00.4576 &            61.1870 &           01.4356 & 47.6849 \\
\Adv        &            68.0645 &              10.1305 &            92.9164 &           04.0835 & 32.7117 & 62.5106 &              11.5933 &            93.1983 &           06.2517 & 38.1014 \\
\DAdv       &            71.8605 &              00.4752 &            94.7193 &           01.3539 & 28.6307 & 64.1210 &              12.8721 &            95.8738 &           03.7152 & 36.1155 \\
\INLP       &            \multicolumn{2}{l}{70.92}     &            \multicolumn{2}{l}{85.36} & 32.5575 & \multicolumn{2}{l}{61.92} &            \multicolumn{2}{l}{91.33} & 39.0584 \\
\FairBatch  &            74.1810 &              00.6203 &            90.4656 &           01.3971 & 27.5232 & 74.1810 &              00.6203 &            90.4656 &           01.3971 & 27.5232 \\
\cmidrule(lr){2-6} \cmidrule(lr){7-11}
\AAdv       &            70.3001 &              00.4699 &            95.1914 &           01.1671 & 30.0867 & 67.3968 &              01.1729 &            96.7237 &           00.6380 & 32.7674 \\
\ADAdv       &           72.3556 &              00.9952 &            95.9129 &           01.5552 & 27.9449 & 64.2336 &              09.1795 &            95.7233 &           02.4305 & 36.0212 \\

\bottomrule
\end{tabular} 
\end{adjustbox}
\caption{\Moji}
\label{table:moji_main_results}
\end{subfigure} \\\hfill

\begin{subfigure}{\textwidth}
\renewrobustcmd{\bfseries}{\fontseries{b}\selectfont}
\centering
\begin{adjustbox}{max width=\linewidth}
\sisetup{
round-mode = places,
}%
\begin{tabular}{
l
S[table-format=2.2, round-precision = 2]@{\,\( \pm \)\,}S[table-format=2.2, round-precision = 2,table-number-alignment = left]
S[table-format=2.2, round-precision = 2]@{\,\( \pm \)\,}S[table-format=2.2, round-precision = 2,table-number-alignment = left]
S[table-format=2.2, round-precision = 2]
S[table-format=2.2, round-precision = 2]@{\,\( \pm \)\,}S[table-format=2.2, round-precision = 2,table-number-alignment = left]
S[table-format=2.2, round-precision = 2]@{\,\( \pm \)\,}S[table-format=2.2, round-precision = 2,table-number-alignment = left]
S[table-format=2.2, round-precision = 2]
}

\toprule
& \multicolumn{5}{c}{\bf Trade-off 5\%} & \multicolumn{5}{c}{\bf Trade-off 10\%} \\
\cmidrule(lr){2-6}\cmidrule(lr){7-11}

\bf Model         & \multicolumn{2}{c}{\bf Accuracy$\uparrow$} &\multicolumn{2}{c}{\bf Fairness$\uparrow$} & \multicolumn{1}{c}{\bf \DTO$\downarrow$ } & \multicolumn{2}{c}{\bf Accuracy$\uparrow$}     &\multicolumn{2}{c}{\bf Fairness$\uparrow$} & \multicolumn{1}{c}{\bf \DTO$\downarrow$ } \\ 
\midrule

\Standard & 81.5181 & 0.2435 & 55.5411 & 2.9533 & 48.1475 & 81.5181 & 0.2435 & 55.5411 & 2.9533 & 48.1475 \\
\Adv &     77.2195 &       2.9662 &     61.9941 &    2.2156 & 44.3102 &    74.5883 &       2.5882 &     62.8701 &    2.8081 & 44.9932 \\
\DAdv       &     78.0985 &       1.3802 &     62.5999 &    1.8781 & 43.3410 &    73.4351 &       3.7638 &     68.7555 &    4.9055 & 41.0111 \\
\INLP       &     \multicolumn{2}{l}{80.41} &     \multicolumn{2}{l}{59.39} & 45.0919 &    \multicolumn{2}{l}{80.41} &     \multicolumn{2}{l}{59.39} & 45.0919 \\
\FairBatch  &     80.5659 &       0.6283 &     61.7986 &    2.0357 & 42.8606 &    80.5659 &       0.6283 &     61.7986 &    2.0357 & 42.8606 \\
\cmidrule(lr){2-6} \cmidrule(lr){7-11}
\AAdv       &     77.6374 &       2.3608 &     63.7065 &    2.5969 & 42.6298 &    73.7561 &       5.1941 &     66.4149 &    5.7328 & 42.6228 \\
\ADAdv      &     77.4654 &       1.6841 &     63.7290 &    3.4463 & 42.7012 &    73.0678 &       2.8756 &     71.6363 &    4.6660 & 39.1133 \\

\bottomrule
\end{tabular} 
\end{adjustbox}
\caption{\Bios}
\label{table:bios_main_results}
\end{subfigure}

\caption{
Evaluation results $\pm$ standard deviation ($\%$) on the test set of sentiment analysis (\Moji) and biography classification (\Bios) tasks, averaged over 5 runs with different random seeds. \DTO\ is measured by the normalized Euclidean distance between each model and the ideal model, and lower is better. Due to the fact that \INLP is a \emph{Post-processing} approach which cannot be run end to end, we only report the its results for 1 run. Least biased models within a given performance tradeoff thresholds are chosen from development set.}
\label{table:main_results}
\end{table*}

We compare our proposed augmented discriminator against the standard discriminator and various competitive baselines: (1) \Standard, which only optimizes performance during training; (2) \Adv~\citep{li-etal-2018-towards}, which adjusts the strength of adversarial training ($\lambda$); (3) \DAdv~\citep{han2021diverse}, which is the SOTA \Adv variation that removes protected information from different aspects, and both $\lambda$ and the strength of orthogonality constraints are tuned; (4) \INLP~\citep{ravfogel-etal-2020-null}, which is a post-processing method, and we tune the number of iteration for null-space projections to control the trade-off, and (5) \FairBatch~\citep{Roh2021FairBatch}, which is the STOA pre-processing method, and we tune adjustment rate for the outer optimization to achieve different trade-offs.

In terms of our own methods, we adopt the Augmentation layer (Figure~\ref{fig:augmentation_layer} for both \Adv and \DAdv, resulting the \textbf{A}ugmented \Adv and \textbf{A}ugmented \DAdv, respectively.
Except for the trade-off hyperparameters, other hyperparameters of each model are the same to \Standard.

Over both datasets, augmented methods achieve better performance--fairness trade-off. I.e., the adversarial method with augmented discriminator achieves better fairness at the same accuracy level, and achieves better accuracy at the same fairness level.
\AAdv, \ADAdv, and \DAdv achieve similar results (their trade-off lines almost overlap), while these methods outperform \Adv consistently.

Moreover, our proposed methods also achieved better trade-offs than other baseline methods: \INLP achieves the worst trade-off over two datasets, as it is a post-process method and the model will not be updated during debiasing; and \FairBatch can achieve a similar trade-off as our proposed methods, but its ability of debiasing is limited due to its resampling strategy. 

\begin{table*}[ht!]

\begin{subfigure}{\textwidth}
\renewrobustcmd{\bfseries}{\fontseries{b}\selectfont}
\centering
\begin{adjustbox}{max width=\linewidth}
\sisetup{
round-mode = places,
}%
\begin{tabular}{
l
S[table-format=2.2, round-precision = 2]@{\,\( \pm \)\,}S[table-format=2.2, round-precision = 2,table-number-alignment = left]
S[table-format=2.2, round-precision = 2]@{\,\( \pm \)\,}S[table-format=2.2, round-precision = 2,table-number-alignment = left]
S[table-format=2.2, round-precision = 2]
S[table-format=2.2, round-precision = 2]@{\,\( \pm \)\,}S[table-format=2.2, round-precision = 2,table-number-alignment = left]
S[table-format=2.2, round-precision = 2]@{\,\( \pm \)\,}S[table-format=2.2, round-precision = 2,table-number-alignment = left]
S[table-format=2.2, round-precision = 2]
}

\toprule
& \multicolumn{5}{c}{\bf Trade-off 5\%} & \multicolumn{5}{c}{\bf Trade-off 10\%} \\
\cmidrule(lr){2-6}\cmidrule(lr){7-11}

\bf Model  & \multicolumn{2}{c}{\bf Accuracy$\uparrow$}     &\multicolumn{2}{c}{\bf Fairness$\uparrow$} & \multicolumn{1}{c}{\bf \DTO$\downarrow$ } & \multicolumn{2}{c}{\bf Accuracy$\uparrow$}     &\multicolumn{2}{c}{\bf Fairness$\uparrow$} & \multicolumn{1}{c}{\bf \DTO$\downarrow$ } \\ 
\midrule

\Adv                 & 68.0645 & 10.1305 & 92.9164 &    4.0835 & 32.7117 & 62.5106 &       11.5933 &     93.1983 &    06.2517 & 38.1014 \\
\AAdv                & 70.3001 & 0.4699 & 95.1914 &    1.1671 & 30.0867 & 67.3968 &       01.1729 &     96.7237 &    00.6380 & 32.7674 \\
\cmidrule(lr){2-6} \cmidrule(lr){7-11}
\Adv+~\textbf{Large} & 72.4706 & 02.2792 & 92.0152 &    3.1307 & 28.6640 & 72.4706 &       02.2792 &     92.0152 &    03.1307 & 28.6640 \\
\cmidrule(lr){2-6} \cmidrule(lr){7-11}
\AAdv+~\textbf{BT}   & 71.3853 & 04.1419 & 90.4159 &    5.4043 & 30.1770 & 66.1765 &       09.8215 &     93.4086 &    04.4786 & 34.4597 \\
\cmidrule(lr){2-6} \cmidrule(lr){7-11}
\Adv+$\yvec$         & 67.8895 & 10.0691 & 93.7331 &    3.8867 & 32.7164 & 67.8895 &       10.0691 &     93.7331 &    03.8867 & 32.7164 \\
\Adv+~\textbf{Sep}   & 68.3046 & 10.3392 & 95.3054 &    3.2996 & 32.0412 & 63.7159 &       12.4834 &     96.2037 &    04.2051 & 36.4821 \\

\bottomrule
\end{tabular} 
\end{adjustbox}
\caption{\Moji}
\label{table:moji_analysis_results}
\end{subfigure} \\\hfill

\begin{subfigure}{\textwidth}
\renewrobustcmd{\bfseries}{\fontseries{b}\selectfont}
\centering
\begin{adjustbox}{max width=\linewidth}
\sisetup{
round-mode = places,
}%
\begin{tabular}{
l
S[table-format=2.2, round-precision = 2]@{\,\( \pm \)\,}S[table-format=2.2, round-precision = 2,table-number-alignment = left]
S[table-format=2.2, round-precision = 2]@{\,\( \pm \)\,}S[table-format=2.2, round-precision = 2,table-number-alignment = left]
S[table-format=2.2, round-precision = 2]
S[table-format=2.2, round-precision = 2]@{\,\( \pm \)\,}S[table-format=2.2, round-precision = 2,table-number-alignment = left]
S[table-format=2.2, round-precision = 2]@{\,\( \pm \)\,}S[table-format=2.2, round-precision = 2,table-number-alignment = left]
S[table-format=2.2, round-precision = 2]
}

\toprule
& \multicolumn{5}{c}{\bf Trade-off 5\%} & \multicolumn{5}{c}{\bf Trade-off 10\%} \\
\cmidrule(lr){2-6}\cmidrule(lr){7-11}

\bf Model  & \multicolumn{2}{c}{\bf Accuracy$\uparrow$} &\multicolumn{2}{c}{\bf Fairness$\uparrow$} & \multicolumn{1}{c}{\bf \DTO$\downarrow$ } & \multicolumn{2}{c}{\bf Accuracy$\uparrow$}     &\multicolumn{2}{c}{\bf Fairness$\uparrow$} & \multicolumn{1}{c}{\bf \DTO$\downarrow$ } \\ 
\midrule

\Adv    & 77.2195 &   2.9662 & 61.9941 &    2.2156 & 44.3102 &  74.5883 &   2.5882 & 62.8701 &    2.8081 & 44.9932 \\
\AAdv   & 77.6374 &   2.3608 & 63.7065 &    2.5969 & 42.6298 &  73.7561 &   5.1941 & 66.4149 &    5.7328 & 42.6228 \\
\cmidrule(lr){2-6} \cmidrule(lr){7-11}
\Adv+~\textbf{Large}  &   77.8762 &   2.8663 & 62.0592 &    1.2736 & 43.9200 & 75.3829 &   5.3457 & 64.0187 &    4.9486 & 43.5965 \\
\cmidrule(lr){2-6} \cmidrule(lr){7-11}
\AAdv+~\textbf{BT}    &   77.3251 &   3.0600 & 63.0041 &    5.0249 & 43.3918 & 74.8598 &   5.1214 & 65.3661 &    6.8683 & 42.7965 \\
\cmidrule(lr){2-6} \cmidrule(lr){7-11}
\Adv+~$\yvec$       &   77.6439 &   1.0993 & 59.8756 &    2.6541 & 45.9322 & 71.9278 &   3.5615 & 64.1095 &    4.5137 & 45.5651 \\
\Adv+~\textbf{Sep}  &   77.5513 &   3.3442 & 65.4962 &    4.6331 & 41.1638 & 72.3909 &   4.8889 & 67.0381 &    6.8553 & 42.9970 \\

\bottomrule
\end{tabular} 
\end{adjustbox}
\caption{\Bios}
\label{table:bios_analysis_results}
\end{subfigure}

\caption{
Evaluation results $\pm$ standard deviation ($\%$) on the test set of sentiment analysis (\Moji) and biography classification (\Bios) tasks, averaged over 5 runs with different random seeds. Least biased models within a given performance tradeoff thresholds are chosen from development set.
\Adv+~\textbf{Large} trains larger discriminators with more parameters than \AAdv as discussed in Section~\ref{sec:LARGE_Discriminator}. \AAdv+~\textbf{BT} employs the instance reweighting for the \AAdv (Section~\ref{sec:Balanced_Discriminator}). \Adv+$\yvec$ and \Adv+~\textbf{Sep} are alternative ways of conditional unlearning as discussed in ~Section~\ref{sec:Alternaative_Discriminator} which takes class labels as inputs and trains class-specific discriminators, respectively.
}
\label{table:analysis_results}
\end{table*}

\begin{table}[t!]
    \centering
    \begin{adjustbox}{max width=\linewidth}
    \begin{tabular}{lrrr}
    \toprule
  \textbf{Model}  &  \Moji  & \Bios \\\hline
   \Adv & 181,202 & 181,202 \\
   \AAdv & 542,402 & 5,057,402   \\
   \Adv+~\textbf{Large}  & 680,450  & 9,013,250  \\
    \bottomrule
    \end{tabular}
    \end{adjustbox}
    \caption{Number of parameters of discriminators over \Moji and \Bios datasets.}
    \label{tab:results_with_parameters}
    \vspace{-1em}
\end{table}

\subsection{Constrained analysis}
\label{sec:constrained_analysis}

As shown in Figure~\ref{fig:tradeoffs}, Pareto curves of different models highly overlap with each other. They can even have multiple intersections, making it hard to say which model strictly outperforms the others.
Alternatively, we follow~\citet{subramanian-etal-2021-evaluating} in comparing the fairest models under a minimum performance constraint, where the fairest models that exceed a performance threshold on the development set are chosen, and evaluated on the test set. 
We provide results under two constrained scenarios --- fairest models on the development set trading off 5\% and 10\% of accuracy.

Table~\ref{table:moji_main_results} shows the results over \Moji dataset. 
With a performance trade-off up to 5\%, both \AAdv and \ADAdv are strictly better than their base models \Adv and \DAdv, respectively.
Also, \ADAdv shows the best fairness than others and consistently outperforms \AAdv.

In terms of other baseline methods, \INLP shows worse performance and fairness than others.
While for \FairBatch, although it shows the smallest \DTO (i.e., best trade-off), its ability of bias mitigation is limited. In both settings, \FairBatch can only achieve a 0.90 fairness score, while \AAdv and \AAdv can achieve a 0.95 fairness score.

For a slack of up to 10\% performance trade-off, augmented adversarial models strictly outperform their base models. 
However, when focusing on the \ADAdv, it can be seen that given a 10\% performance trade-off, the fairness score is even not as good as that at the 5\% level. 
The less fairness is caused by the inconsistency between the results over the development set and test set, i.e., the fairest model is selected based on the development set but does not generalize well on the test set. 

For the occupation classification, results are summarized in Table~\ref{table:bios_main_results}. 
Similarly, our proposed methods consistently outperform others for both 5\% and 10\% performance settings, confirming the conclusion based on trade-off plots that adding augmentation layers for adversarial debiasing leads to better fairness.

\section{Analysis}
To better understand the effectiveness of our proposed methods, we perform three sets of experiments: (1) analysis of the effects of extra parameters in the Augmentation layer, (2) an ablation study of the effect of balanced discriminator training, and (3) a comparison with alternative ways of incorporating target labels in adversarial training. 
Similar to constrained analysis, we report the fairest models under a minimum performance constraint in Table~\ref{table:analysis_results}. 

\subsection{Affects of extra parameters}
\label{sec:LARGE_Discriminator}

Augmentation layers increase the number of parameters in discriminators, and thus an important question is whether the gains of augmentation layers are because of additional parameters.

We conduct experiments with two scenarios by controlling the number of parameters of a discriminator in \Adv, namely \Adv+~\textbf{Large}, and compare it with \AAdv models.
Specifically, we employ a larger discriminator with an additional layer and more hidden units within each layer, leading to roughly the same number of parameters as the augmented discriminator.

Table~\ref{tab:results_with_parameters} summaries the number of parameters of discriminators under different settings. 
For \Adv, the discriminators are 2-layer MLP. Thus its parameter size is only affected by its architecture, resulting in the same amount for two datasets. While for \AAdv, the amount of class-specific components within the augmentation layer is determined by the total distinct target classes, which are 2 and 28 for \Moji and \Bios, respectively, leading to a much larger discriminator for the \Bios dataset.

As shown in Table~\ref{table:analysis_results}, although \Adv+~\textbf{Large} contains more parameters than others, it is not as good as \AAdv. 
Over the \Moji dataset, \Adv+~\textbf{Large} achieves a better trade-off, however, the best fairness that can be achieved is limited to 0.92, which is similar to the \FairBatch method.
Similarly, for \Bios (Table~\ref{table:bios_analysis_results}), \Adv+~\textbf{Large} leads to less fair models, and worse trade-offs than \AAdv.

\subsection{Balanced discriminator training}
\label{sec:Balanced_Discriminator}

One problem is that the natural distribution of the demographic labels is imbalanced. E.g. \ in \Bios, 87\% of nurses are female while 90\% of surgeons are male.
A common way of dealing with this label imbalance is instance reweighting~\citep{han2021balancing}, which reweights each instance inversely proportional to the frequency of its demographic label within its target class when training the discriminators.

We experiment with the balanced version of discriminator training to explore the impact of such balanced training methods.
The instance weighting is only applied to the discriminator training, while during unlearning, instances are assigned identical weights to reflect the actual proportions of protected groups.

\AAdv methods with balanced training are denoted as \AAdv+~\textbf{BT}, and corresponding results over two datasets are shown in~Table~\ref{table:analysis_results}.
In contrast to our intuition that balanced training enhances the ability to identify protected attributes and leads to better bias mitigation, the \AAdv+~\textbf{BT} methods can not improve the trade-offs in any setting.
This might be because of the imbalanced training of the main task model, which mainly focuses on the majority group within each class, in which case the imbalanced trained discriminator is also able to identify protected information properly.

\subsection{Alternative ways of conditional unlearning}
\label{sec:Alternaative_Discriminator}

Essentially, our method is one way of unlearning protected information conditioned on the class labels. 
Alternatively, we can concatenate the hidden representations $\hvec$ and target labels $\yvec$ as the inputs to the discriminator~\citep{wadsworth2018achieving}, as well as train a set of discriminators, one for each target class~\citep{zhao2019conditional}.

Taking target labels $\yvec$ as the inputs (\Adv+$\yvec$) can only capture the target class conditions as a different basis, which is insufficient. As such, it can be seen from Table~\ref{table:analysis_results} that, \Adv+$\yvec$ only leads to minimum improvements over the \Adv, but is not as good as \AAdv.

The other alternative method, \Adv+~\textbf{Sep} which trains one discriminator for each target class, is the closed method to our proposed augmentation layers. 
However, \Adv+~\textbf{Sep} has two main limitations: (1) discriminators are trained separately, which cannot learn the shared information across different class like the shared component of augmentation layers, and (2) the training and unlearning process for multiple discriminators are much complicated, especially when combined with other variants of adversarial training, such as Ensemble Adv~\citep{elazar2018adversarial} and \DAdv~\citep{han2021diverse} that employs multiple sub-discriminators.

As shown in Table~\ref{table:analysis_results}, although \Adv+~\textbf{Sep} shows improvements to \Adv+$\yvec$, \AAdv still outperforms \Adv+~\textbf{Sep} significantly (1 percentage point better in in terms of trade-off on average).

\section{Conclusion}
We introduce an augmented discriminator for adversarial debiasing. 
We conducted experiments over a binary tweet sentiment analysis with binary author race attribute and a multiclass biography classification with the multiclass protected attribute.
Results showed that our proposed method, considering the target label, can more accurately identify protected information and thus achieves better performance--fairness trade-off than the standard adversarial training.

\section*{Ethical Considerations}
This work aims to advance research on bias mitigation in NLP. Although our proposed method requires access to training datasets with protected attributes, this is the same data assumption made by adversarial training. Our target is to remove protected information during training better. To avoid harm and be trustworthy, we only use attributes that the user has self-identified for experiments. Moreover, once being trained, our proposed method can make fairer predictions without the requirement of demographic information. All data in this study is publicly available and used under strict ethical guidelines.

\bibliography{custom}
\bibliographystyle{acl_natbib}

\clearpage

\appendix

\section{Dataset}
\subsection{\Moji}
We use the train, dev, and test splits from \citet{han2021diverse} of 100k/8k/8k instances, respectively. 
This training dataset has been artificially balanced according to demographic and task labels, but artificially skewed in terms of race--sentiment combinations, as follows: \AAE--\happy = 40\%, \SAE--\happy = 10\%, \AAE--\sad = 10\%, and \SAE--\sad = 40\%.

\subsection{\Bios}
Since the data is not directly available, in order to construct the dataset, we use the scraping scripts of \citet{ravfogel-etal-2020-null}, leading to a dataset with 396k biographies.\footnote{There are slight discrepancies in the dataset composition due to data attrition: the original dataset~\citep{de2019bias} had 399k instances, while 393k were collected by~\citet{ravfogel-etal-2020-null}.}
Following~\citet{ravfogel-etal-2020-null}, we randomly split the dataset into train (65\%), dev (10\%), and test (25\%).

Table~\ref{table:bios_distribution} shows the target label distribution and protected attribute distribution.

\begin{table}[t!]
\renewrobustcmd{\bfseries}{\fontseries{b}\selectfont}
\centering
\begin{adjustbox}{max width=\linewidth}
\sisetup{
round-mode = places,
}%
\begin{tabular}{
l
r
S[table-format=1.3, round-precision = 3]
S[table-format=1.3, round-precision = 3]
S[table-format=1.3, round-precision = 3]
S[table-format=1.3, round-precision = 3]
}

\toprule
       Profession &  Total &  \multicolumn{1}{c}{male\_rest} &  \multicolumn{1}{c}{male\_wealthy} &  \multicolumn{1}{c}{female\_rest} &  \multicolumn{1}{c}{female\_wealthy} \\
\midrule
        professor &  21715 &   0.091964 &      0.461708 &     0.072715 &        0.373613 \\
        physician &   7581 &   0.084422 &      0.424482 &     0.079937 &        0.411159 \\
         attorney &   6011 &   0.099152 &      0.511728 &     0.062219 &        0.326901 \\
     photographer &   4398 &   0.110505 &      0.530923 &     0.055935 &        0.302638 \\
       journalist &   3676 &   0.093308 &      0.407236 &     0.085691 &        0.413765 \\
            nurse &   3510 &   0.011396 &      0.075499 &     0.149288 &        0.763818 \\
     psychologist &   3280 &   0.064939 &      0.307012 &     0.105488 &        0.522561 \\
          teacher &   2946 &   0.061439 &      0.351324 &     0.095044 &        0.492193 \\
          dentist &   2682 &   0.113348 &      0.520880 &     0.063013 &        0.302759 \\
          surgeon &   2465 &   0.123732 &      0.726572 &     0.023935 &        0.125761 \\
        architect &   1891 &   0.116341 &      0.641460 &     0.034373 &        0.207827 \\
          painter &   1408 &   0.088778 &      0.473011 &     0.075284 &        0.362926 \\
            model &   1362 &   0.024963 &      0.149046 &     0.129956 &        0.696035 \\
             poet &   1295 &   0.073359 &      0.459459 &     0.081853 &        0.385328 \\
software\_engineer &   1289 &   0.137316 &      0.697440 &     0.024825 &        0.140419 \\
        filmmaker &   1225 &   0.096327 &      0.555918 &     0.058776 &        0.288980 \\
         composer &   1045 &   0.141627 &      0.704306 &     0.017225 &        0.136842 \\
       accountant &   1012 &   0.094862 &      0.553360 &     0.063241 &        0.288538 \\
        dietitian &    730 &   0.012329 &      0.050685 &     0.120548 &        0.816438 \\
         comedian &    499 &   0.090180 &      0.693387 &     0.030060 &        0.186373 \\
     chiropractor &    474 &   0.143460 &      0.618143 &     0.031646 &        0.206751 \\
           pastor &    453 &   0.145695 &      0.593819 &     0.035320 &        0.225166 \\
        paralegal &    330 &   0.027273 &      0.124242 &     0.148485 &        0.700000 \\
     yoga\_teacher &    305 &   0.029508 &      0.134426 &     0.121311 &        0.714754 \\
interior\_designer &    267 &   0.041199 &      0.164794 &     0.123596 &        0.670412 \\
 personal\_trainer &    264 &   0.098485 &      0.412879 &     0.068182 &        0.420455 \\
               dj &    244 &   0.155738 &      0.709016 &     0.024590 &        0.110656 \\
           rapper &    221 &   0.153846 &      0.746606 &     0.009050 &        0.090498 \\
\hline
\textbf{Total} & 72578 & 0.08863567 & 0.45079776 & 0.07463694 & 0.38592962\\
\bottomrule
\end{tabular} 
\end{adjustbox}
\caption{Training set distribution of the \Bios dataset.}

\label{table:bios_distribution}
\end{table}

\section{Reproducibility}
\label{sec:Reproducibility}

\subsection{Computing infrastructure}
We conduct all our experiments on a Windows server with a 16-core CPU (AMD Ryzen Threadripper PRO 3955WX), two NVIDIA GeForce RTX 3090s with NVLink, and 256GB RAM.

\subsection{Computational budget}
Over the \Moji dataset, we run experiments with 108 different hyperparameter combinations (each for 5 runs with different random seeds) in total, which takes around 300 GPU hours in total and 0.56 hrs for each run.
Over the \Bios dataset, we run experiments with 162 different hyperparameter combinations for around 466 GPU hours and 0.58 hrs for each run.

\subsection{Model architecture and size}
In this paper, we used pretrained models as fixed encoder, and the number of fixed parameters of DeepMoji~\citep{felbo2017} for \Moji and uncased BERT-base~\citep{devlin2019bert} for \Bios are approximately 
22M and 110M, resp.
The number of remaining trainable parameters of the main model is about 1M for both tasks.

As for the standard discriminator, we follow~\cite{han-etal-2021-decoupling} and use the same architecture for both tasks, leading to a 3-layer MLP classifier with around 144k parameters.

\subsection{Hyperparameters}
For each dataset, all main task model models in this paper share the same hyperparameters as the standard model.
Hyperparameters are tuned using grid-search, in order to maximize accuracy for the standard model.
Table~\ref{tab:standand_hyperparameter} summaries search space and best assignments of key hyperparameters.

\begin{table*}[ht!]
\centering
    \begin{adjustbox}{max width=\linewidth}

    \centering
    \begin{tabular}{cccc}
        \toprule 
        & & \multicolumn{2}{c}{\bf Best assignment} \\
        \cmidrule(lr){1-2} \cmidrule(lr){3-4}
        \textbf{Hyperparameter} & \textbf{Search space} & \bf \Moji & \bf \Bios\\
        \cmidrule(lr){1-2} \cmidrule(lr){3-4}
        number of epochs & - & \multicolumn{2}{c}{100}\\
        \cmidrule(lr){1-2} \cmidrule(lr){3-4}
        patience & - & \multicolumn{2}{c}{10}\\
        \cmidrule(lr){1-2} \cmidrule(lr){3-4}
        embedding size & - & 2304 & 768\\
        \cmidrule(lr){1-2} \cmidrule(lr){3-4}
        hidden size & - & \multicolumn{2}{c}{300}\\
        \cmidrule(lr){1-2} \cmidrule(lr){3-4}
        number of hidden layers & \emph{choice-integer}[1, 3] & \multicolumn{2}{c}{2}\\
        \cmidrule(lr){1-2} \cmidrule(lr){3-4}
        batch size & \emph{loguniform-integer}[64, 2048] & 1024 & 512 \\
        \cmidrule(lr){1-2} \cmidrule(lr){3-4}
        output dropout & \emph{uniform-float}[0, 0.5] & 0.5 & 0.3 \\
        \cmidrule(lr){1-2} \cmidrule(lr){3-4}
        optimizer & - & \multicolumn{2}{c}{Adam~\citep{kingma:adam}}\\
        \cmidrule(lr){1-2} \cmidrule(lr){3-4}
        learning rate  & \emph{loguniform-float}[$10^{-6}$, $10^{-1}$] & $3\times10^{-3}$ & $10^{-3}$ \\
        \cmidrule(lr){1-2} \cmidrule(lr){3-4}
        \textbf{l}earning \textbf{r}ate \textbf{s}cheduler & - & \multicolumn{2}{c}{reduce on plateau}\\
        \cmidrule(lr){1-2} \cmidrule(lr){3-4}
        \textbf{LRS} patience & - & \multicolumn{2}{c}{2 epochs}\\
        \cmidrule(lr){1-2} \cmidrule(lr){3-4}
        \textbf{LRS} reduction factor & - & \multicolumn{2}{c}{0.5}\\
        \bottomrule
    \end{tabular}
    \end{adjustbox}
    \caption{Search space and best assignments.}
    \label{tab:standand_hyperparameter}
\end{table*}

To explore trade-offs of our proposed method at different levels, we tune $\lambda$ log-uniformly to get a series of candidate models.
Specifically, the search space of $\lambda$ with respect to \Moji and \Bios are both \emph{loguniform-float}[$10^{-3}$, $10^{3}$].

\end{document}